\documentclass{article}

\usepackage{arxiv}

\usepackage[utf8]{inputenc} % allow utf-8 input
\usepackage[T1]{fontenc}    % use 8-bit T1 fonts
\usepackage{hyperref}       % hyperlinks
\usepackage{url}            % simple URL typesetting
\usepackage{booktabs}       % professional-quality tables
\usepackage{amsfonts}       % blackboard math symbols
\usepackage{nicefrac}       % compact symbols for 1/2, etc.
\usepackage{microtype}      % microtypography
\usepackage{lipsum}		% Can be removed after putting your text content
\usepackage{graphicx}
\usepackage[super,sort,comma]{natbib}
\usepackage{doi}

\title{Direct 3D Printing of Soft Fluidic Actuators with Graded Porosity}

\author{ \href{}{\hspace{1mm} Nick~Willemstein}\\
	Department of Biomechanical Engineering\\
	University of Twente\\
	Enschede, The Netherlands \\
	\texttt{n.willemstein@utwente.nl} \\
	\And
	{\hspace{1mm}Herman~van der Kooij} \\
	Department of Biomechanical Engineering\\
	University of Twente\\
	Enschede, The Netherlands \\
	\texttt{h.vanderkooij@utwente.nl} \\
	\And
	{\hspace{1mm}Ali~Sadeghi} \\
	Department of Biomechanical Engineering\\
	University of Twente\\
	Enschede, The Netherlands \\
	\texttt{a.sadeghi@utwente.nl} \\}

\renewcommand{\headeright}{}
\renewcommand{\undertitle}{}
\renewcommand{\shorttitle}{\textit{Direct 3D Printing of Soft Fluidic Actuators with Graded Porosity}}

\begin{document}
\maketitle

\begin{abstract}
New additive manufacturing methods are needed to realize more complex soft robots. One example is soft fluidic robotics, which exploits fluidic power and stiffness gradients. Porous structures are an interesting type for this approach, as they are flexible and allow for fluid transport. Within this work, the Infill-Foam (InFoam) is proposed to print structures with graded porosity by liquid rope coiling (LRC). By exploiting LRC, the InFoam method could exploit the repeatable coiling patterns to print structures. To this end, only the characterization of the relation between nozzle height and coil radius and the extruded length were necessary (at a fixed temperature). Then by adjusting the nozzle height and/or extrusion speed the porosity of the printed structure could be set. The InFoam method was demonstrated by printing porous structures using styrene-ethylene-butylene-styrene (SEBS) with porosities ranging from 46\% to 89\%. In compression tests, the cubes showed large changes in modulus (more than 200 times), density (-89\% compared to bulk), and energy dissipation.
The InFoam method combined coiling and normal plotting to realize a large range of porosity gradients. This grading was exploited to realize rectangular structures with varying deformation patterns, which included twisting, contraction, and bending. Furthermore, the InFoam method was shown to be capable of programming the behavior of bending actuators by varying the porosity. Both the output force and stroke showed correlations similar to those of the cubes. Thus, the InFoam method can fabricate and program the mechanical behavior of a soft fluidic (porous) actuator by grading porosity.
\end{abstract}

\section{Introduction}

Soft robotics utilize softness and smart design to realize robots that are inherently safe for human-robot interaction and compliant to unforeseen disturbances. Soft fluidic robotics are among the most popular type due to their flexibility, wide range of applications, and (relative) simplicity for actuation. They have been used in a broad range of applications from grippers\cite{mosadegh2014pneumatic} to autonomous systems\cite{wehner2016integrated}.

Soft fluidic robotics work by transferring fluidic power (flow and pressure) from one place to another. This power can be used for actuation\cite{mosadegh2014pneumatic}, sensing\cite{cheng2018soft}, and control\cite{wehner2016integrated,mahon2019soft}. Soft fluidic actuators (SFAs) consist of mechanically compliant chambers that can be deformed using fluidic power. The deformation of these chambers is defined by the stiffness gradient of the structure. This gradient allows for motions such as translation, bending, and twisting \cite{connolly2015mechanical}. 
Researchers employ a broad number of methods to realize this stiffness gradient, which includes exploiting material properties and/or geometry. An approach is to add a second material such as fibers\cite{connolly2015mechanical} textile\cite{mosadegh2014pneumatic}, and/or paper\cite{martinez2012elastomeric} in a dedicated geometry. Other approaches use a single material and exploit geometrical features, such as origami\cite{paez2016design} and kirigami\cite{dias2017kirigami}. A prominent example of exploiting internal geometry is porous structures, such as foam.

An advantage of porous structures, is that they can, inherently, allow for fluidic transport and are flexible. In addition, porous structures are compressible. This feature has been exploited to realize vacuum-driven actuators. Examples thereof include: a continuum soft robot\cite{robertson2017new}, high-force sensorized foam actuators\cite{murali2021sensorized}, and bending actuators\cite{yamada2019laminated}. Lastly, positive pressure actuation is also feasible with porous structures for both actuation\cite{mac2015poroelastic} and control\cite{futran2018leveraging}.

Existing methods for fabricating porous structures span a broad spectrum of approaches. A popular approach is to use commercial foams combined with semi-automatic/manual processing, such as laser-cutting, gluing, and coating with silicone or laminating (for airtightness)\cite{robertson2017new,murali2021sensorized,yamada2020actuatable}. Another method is to use casting with a sacrificial scaffold/material such as salt \cite{mac2015poroelastic,futran2018leveraging}, sugar\cite{bilent2019influence} or Polyvinyl Alcohol (PVA)\cite{bilent2020fabrication} that can be removed afterwards. 

To realize the multifunctional nature of soft robotic systems (including SFAs based on porous structures), additive manufacturing (AM) is considered a promising fabrication method. Literature has already demonstrated that AM can realize (sensorized) SFAs\cite{morrow2016directly,yap2016high,walker2019zero,khondoker2019direct,georgopoulou2021sensorized}. However, the fabrication of porous structures for soft robots by AM methods is less developed. 

Recently, a custom ink\cite{yirmibesoglu2021multi} for realizing local porosity was developed by adding a porogen. By regulating the ratio of ammonium bicarbonate (the porogen) and silicone rubber the porosity can be changed locally. This approach requires modification of the material itself.

Besides using a chemical reaction, a porous structure can also be realized by the fabrication of miniature patterns during the AM process. Liquid rope coiling (LRC) is an interesting candidate for the fabrication of these patterns. LRC is the coiling of a viscous fluid, such as honey, due to buckling when falling from an elevated height\cite{ribe2004coiling}. 

\begin{figure*}[t]
 	\centering
 	\includegraphics[width=0.8\textwidth]{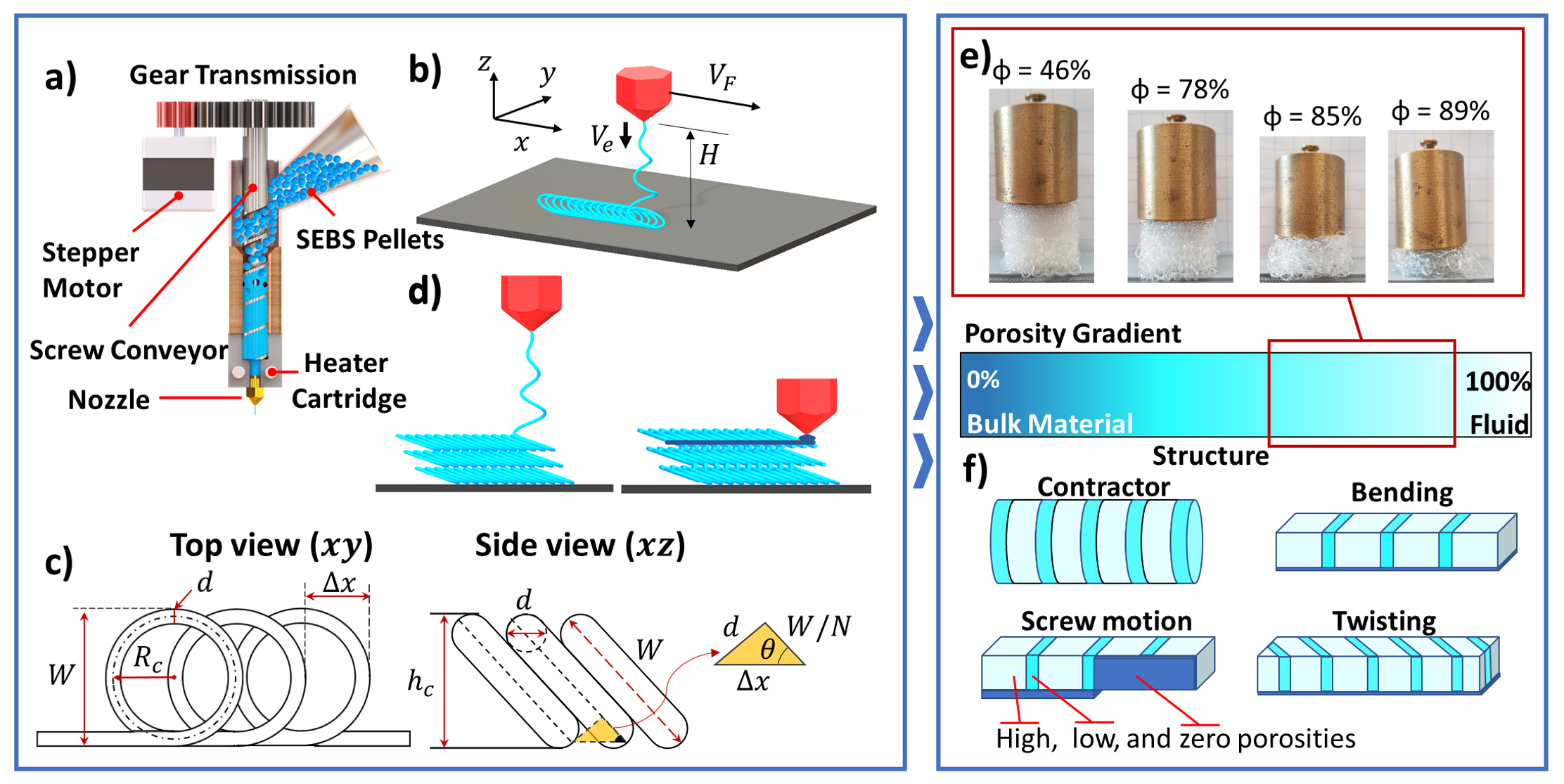}
 	\caption{(a) Screw extruder, (b) liquid rope coiling setup with important parameters, (c) coil geometry, (d) grading approach, (e) cubes with different porosity under the same load, and (f) porosity gradients for soft vacuum actuators.}
 	\label{fig:lrc_intro}
 \end{figure*}

LRC has already been exploited to fabricate porous structures\cite{lipton20163d,brun2017molten,tian2017silicone,zou2020spiderweb,emery2021applied}. Results in the literature indicate that LRC can realize stiffness changes of more than one order of magnitude\cite{lipton20163d,emery2021applied}. In addition, it has been shown to be feasible to print different patterns by proper selection of process parameters. \cite{tian2017silicone,brun2017molten,emery2021applied} Such an approach has been applied to AM processes extrude molten materials that solidify in the process, such as glass\cite{brun2017molten} and thermoplastics\cite{zou2020spiderweb,emery2021applied}.
Although these initial results indicate that porous structures can be realized with LRC and AM, its application to soft robotics has, to the knowledge of the authors, not yet been explored. In addition, we also explore how to use LRC to create multiple levels of porosity in a single structure to realize mechanical programming with porosity.
Within this work, a new printing approach that exploits LRC for the fabrication of porous structures using hyperelastic materials is proposed. This new approach is referred to as the Infill Foam (InFoam) approach. We demonstrate how LRC's coiling behavior can be incorporated into the fused deposition modeling (FDM) process to fabricate porous hyperelastic structures. To use hyperelastic materials within the FDM process a screw extruder (see Fig. \ref{fig:lrc_intro}(a)) was used\cite{saari2015additive,khondoker2019direct}. 
Subsequently, the InFoam method is used for mechanical programming of stiffness, density, and energy dissipation. Then the extension to graded porosity is demonstrated as a tool for programming the deformation and to program the behavior of soft bending vacuum actuators.

%%%%%%%%%%%%%%%%%%%%
\section{Materials and Methods}
%%%%%
\subsection{From Coiling to Printing}
Liquid rope coiling is a phenomenon that can be observed when extruding a viscous liquid above the printing (see  Fig. \ref{fig:lrc_intro}(b)). This situation is also observed in extrusion-based AM methods such as FDM. When there is a gap between the nozzle and printing surface, the deposited material acts as a flexible (liquid) rope that buckles, which leads to a pattern of coils on the surface (see Fig. \ref{fig:lrc_intro}(b)). A wide variety of shapes can be achieved through LRC, such as a meander and figure of eight\cite{chiu2006fall,yuk2018new}. Within this work, the focus is purely on the circular coil pattern. The repetitive nature of the circular coiling pattern makes it an interesting tool for 3D printing.

Literature has shown that the shape of the deposited coils (called coiling pattern) depends on three parameters, namely: (I) the viscosity of the extruded material, (II) the height ($H$) of the nozzle above the platform, (III) the ratio of extrusion ($V_e$) and printhead speed ($V_F$)\cite{chiu2006fall,ribe2012liquid}. All three parameters can be set by the user in the FDM process. The viscosity can (primarily) be set using the printing temperature and the height by moving the nozzle. Whereas the ratio of extrusion and printhead speed is set by the extrusion multiplier ($\alpha$). The variable $\alpha$ is the screw rotation (in radians) per millimeter moved of the nozzle. In essence, $\alpha$ scales the distance traveled by the printhead to compute the required screw rotation, which defines the ratio of speeds.

These parameters can be set to control the coiling pattern and to plan the desired shape. However, modeling the coiling pattern is not straightforward due to the changes in material properties (such as viscosity) due to the solidification of the thermoplastic during deposition. Within this work, an experimental method, similar to\cite{brun2017molten}, is used to characterize the coiling pattern. 

Similar to conventional printing, the InFoam method needs the dimensions (width and height) of the coiling pattern for path planning. The coiling pattern consists of coils that stack on top of each other in line with the movement of the printhead (see Fig. \ref{fig:lrc_intro}(b)). The geometry of this coiling pattern can be used to compute the width and height. As observed in Fig. \ref{fig:lrc_intro}(c), the width $W$ of the coiling pattern is equal to the diameter of the coil, which is equal to
\begin{equation}
W = 2R_c+d
\label{eq:Rc}
\end{equation}
Within this equation, $R_c$ is the coil radius and $d$ the nozzle diameter. 
To compute the coil height $h_c$, a second parameter of the coiling pattern needs to be quantified, namely the $N$-value. The $N$-value represents the coil density within a row of coils, which is defined by the number of coils within one (outer) coil diameter ($W$). This value can be computed after deposition by measuring the distance between two coils $\Delta x$ and using the relation:  
\begin{equation}
N=\sqrt{\frac{W^2}{\Delta x^2-d^2}}
\label{eq:N1}
\end{equation}
The $N$-value can also be computed from the coiling behavior by examining the ratio of the extruded length over the linear distance of one outer coil diameter ($W$). This modelling approach allows for a relation between the (user set) parameters $\alpha$ and $H$ (controls the value of $R_c$) and the $N$-value. After rewriting the coil density ($N$-value) can be expressed as (see derivation in Supplementary Information)
\begin{equation}
	N = \frac{G\alpha W}{2 \pi R_c}
\label{eq:N}
\end{equation} 
The variable $G$ (mm/rad) represents the length of material extruded per radian turned by the screw. The exact value of $G$ is defined by the material and geometry of the screw and barrel\cite{crawford2020plastics}. Within this work, this value is acquired by fitting experimental data. Thus, in Equation \ref{eq:N} moving the linear distance $W$ the screw rotates $\alpha W$ radians leading to an extruded length of $G\alpha W$. This length is used to compute the coil density by dividing it by the circumference of the coil ($2\pi R_c$).  Thus, the coil density can be adjusted by changing the extrusion multiplier ($\alpha$) under the assumption of constant $R_c$. Adjustment of the extrusion multiplier for the desired density at a different $R_c$ requires scaling by the factor $W/(2\pi R_c)$. The scaled extrusion multiplier $\alpha W/(2 \pi R_c)$ is referred to as the coil density multiplier.

The coil density and radius can be used to compute the coil height $h_c$. The coil can be modelled as a rectangle with rounded corners (see Fig. \ref{fig:lrc_intro}(c)) under an angle of $\theta=\arctan\left(\frac{Nd}{W}\right)$. The height is then equal to
\begin{equation}
    h_c = W \sin\left(\theta\right)+\left(1-\sin\left(\theta\right)\right)d
    \label{eq:hc}
\end{equation}

The preceding equations can be used for path planning and characterization of the coiling pattern. Specifically, Equations \ref{eq:Rc} and \ref{eq:hc} can be used to compute the width and height of the coiling pattern. Whereas Equation \ref{eq:N1} and \ref{eq:N} can be used to measure and set the desired coil density ($N$-value), respectively. To use these equations the value of $R_c$ needs to be characterized to relate it to $H$. 
The geometry of the deposited coiling patterns will introduce porosity. This porosity can be used for mechanical programming of a structure\cite{gibson1982mechanics}. The possibility of grading the porosity allows for localized mechanical programming. Combining LRC and the normal plotting of the 3D printer enables such functionality by using the deposited coils as a scaffold for the normal plotting. A solution for printing non-porous zones is to move the nozzle down after extruding a layer of coils (i.e. the scaffold) and then locally plot additional material(see Fig. \ref{fig:lrc_intro}(d)). This simple approach enables porosity grading and ensures proper force transfer by welding layers of coils together. 

%%%%%
\subsection{Mechanical Programming with Porosity}
The InFoam method can produce (graded) coiling patterns with different coiling radii and $N$-values by adjusting the extrusion multiplier ($\alpha$) and/or the height ($H$). The realized coiling patterns can be used to print porous structures. The porosity will have a major influence on the resulting mechanical properties\cite{gibson1982mechanics}. The repetitive nature of the coiling pattern was exploited to estimate the porosity $\phi$ (in percent), which is equal to (see Supplementary Information for the derivation)
\begin{equation}
	\phi = 100\left(1-\frac{G \alpha \pi d^2}{ 4Wh_c}\right) 
	\label{eq:Gman}
\end{equation} 
This equation estimates the porosity by dividing the extruded volume (numerator) by the spanned volume (denominator). The spanned volume is based on a rectangular bar of coil height $h_c$ and outer coil diameter $W$ whereas the length is canceled out (see Supplementary Information). This equation shows that the user can decrease the porosity nonlinearly by both $\alpha$ and $H$. The latter is implicit as changing the $H$ will change the coil radius $R_c$, which is related to $W$ and $h_c$.   

The porosity can be used to program a wide variety of mechanical properties, which include elastic modulus and density. The elastic modulus decreases quadratically with the porosity\cite{gibson1982mechanics}. The large change in stiffness is visualized by the samples shown in Fig. \ref{fig:lrc_intro}(e) wherein the same load lead to deformations ranging from minor ($\phi=46$\%) to large ($89$\%). 

In general, the effect of porosity on mechanical properties can be captured in a power-law relation through the phenomenological model\cite{gibson1982mechanics}:
\begin{equation}
	\frac{p_p}{p_s} = C\left(1-\phi/100\right)^n 
\end{equation}
Wherein the $p_p$ and $p_s$ represent the mechanical property of interest of the porous structure and solid, respectively. Whereas $C$ and $n$ are two fitting values. The relation between density and porosity for air (and other low-density fluids), is an example of such a relation (with $n=1$ and $C=1$)
\begin{equation}
 \frac{\rho_p}{\rho_b}= 1-\phi/100
	\label{eq:dens}
\end{equation}
The variables $\rho_p$ and $\rho_b$ represent the densities of the porous structure and bulk material, respectively. This equation approximates the porosity well as low-density fluids (such as air) have densities that are multiple orders of magnitude below that of the solid material.

%%
%%%%%%%%%%%%%%%%%%%%%%%%%%%%%%%%%%%%%%%%%%%%%%%
%%%%%%%%%%%%%%%%%%%%%%%%%%%%%%%%%%%%%%%%%%%%%%%
\subsection{Experimental Setup \& Methodology}
\subsubsection{3D Printing Setup}
For printing with the InFoam method, a modified Creality Ender 5 Plus (Shenzhen Creality 3D Technology Co., Ltd., China) was used (see Figure S1 in the Supplementary Information). The printer was modified to use a screw extruder (see schematic in Fig. \ref{fig:lrc_intro}(a)) to allow for the use of thermoplastic elastomers. To magnify the torque a 22:63 gear ratio between the stepper motor and screw was added. However, the extrusion multiplier ($\alpha$) will always refer to the screw's rotation. As printing material G1657 styrene-ethylene-butylene-styrene (SEBS) pellets (Kraton Corporation, USA) with a Shore hardness of 47A and bulk density of 900 kg/m$^3$ were used.  In addition, all experiments were conducted with a printing temperature of 230$^\circ$C and with a nozzle diameter of $d=0.4$ mm. Lastly, the G-code for printing with the InFoam method was generated using a custom script in MATLAB (The Mathworks, Inc., USA). Printing with the InFoam method is shown in Supplementary Movie 1 in the Supplementary Information.

%%%%%%%%%%%%%%%%%%%%%%%%%%%%%%%%%%%%%%%%%%%%%%%    
\subsubsection{Characterization of Coiling Pattern}
Using the InFoam method for printing complex structures requires characterization of the coiling pattern (i.e. the coil radius ($R_c$) and density ($N$-value)). The LRC effect indicates that the coil radius and the $N$-value are controlled by the viscosity, extrusion multiplier ($\alpha$), and height ($H$). Within this experiment, the $H$ and $\alpha$ were varied at a fixed temperature but due to the shear-thinning behavior of the thermoplastic, the extrusion speed will also affect the viscosity (and therefore the coil radius).

For this characterization, several lines were printed. Afterwards, the printed lines were scanned with a USB camera, and the acquired images were analyzed using ImageJ\cite{schindelin2012fiji}. To relate the pixel distance to real-world distances dots on the printbed were used as a reference distance. Specifically, the width ($W$) and linear distance ($\Delta x$) were measured. These measurements were used to determine $R_c$ and the $N$-value using Equations \ref{eq:Rc} and \ref{eq:N1}, respectively.

%%%%%%%%%%%%%%%%%%%%%%%%%%%%%%%%%%%%%%%%%%%%%%%
%%%%%%%%%%%%%%%%%%%%%%%%%%%%%%%%%%%%%%%%%%%%%%%
\subsubsection{Mechanical Programming of Soft Cubes}
To investigate the effect of the coil radius ($R_c$) and density ($N$-value) on the porosity, stiffness, and damping a set of 25x25x25 mm$^3$ cubes were printed.   The cubes were printed (in triplicate) for different heights ($H$) and $N$-values. The corresponding extrusion multiplier ($\alpha$) and $R_c$ were based on the results of the coiling characterization to use for process planning (i.e Equations \ref{eq:Rc}, \ref{eq:N}, and \ref{eq:hc}). The cubes were divided into two groups of six (with one the same in both $[H,N]=[6,3]$) with either a change in $N$-value$=[2.2,3,6.3,8.5,10.75,12.75]$ or $H=[2,4,6,8,10,15]$. All cubes were printed with the coils in the same direction for each layer.

After printing the cubes, the mass and height were measured using a scale and caliper, respectively. The resulting porosity was then computed using the measured density by dividing the mass by the cube's volume. The resulting porosity (in percent) is then computed using Equation \ref{eq:dens}. 

Subsequently, the stiffness and damping were investigated using a universal tensile tester (Instron 3343, Instron, USA). The testing was done in four phases. Firstly, the cubes were pre-loaded by compressing to 5 mm at a speed of 5 mm/min. Secondly, it returned to its original position at 60 mm/min. These two load steps were performed to have a similar pre-load for all cubes. In the third phase, the cubes were compressed for 21 mm at 10 mm/min. This loading step was slow to have a primarily elastic response. Lastly,  the tensile tester returned to its original position with a speed of 60 mm/min. The speed was higher in this step to increase the viscous damping component. These loading steps were performed for all three axes (each in triplicate).

The stiffness was approximated by computing the segment modulus at different levels of strain. A quadratic curve was fitted at specific levels of strain using the twelve preceding points. The result of this fitting was used to compute the segment modulus.
In addition, the dissipated energy by hysteresis was investigated to evaluate the damping. The dissipated energy was computed by taking the ratio of the stress-strain curve's inner shape (i.e. between the forward and backward motion) and the overall curve.

%%%%%%%%%%%%%%%%%%%%%%%%%%%%%%%%%%%%%%%%%%%%%%%  
\subsubsection{Graded Porosity for Soft Fluidic Actuators}
By moving from a single level of porosity to a graded porosity, the deformation can be programmed. To demonstrate this grading capability, a set of soft vacuum actuators with different porosity gradients were printed. These included the contraction pattern of Fig. \ref{fig:lrc_intro}(f), which used high (80+\%) and low porosity discs stacked on top of each other to realize contraction. The low porosity discs had a low porosity ring (<20\%) on the outside and high porosity inside to improve contraction but preserve airflow (the exceptions are the top and bottom parts, which are completely low porosity). In addition,bending actuators were printed (see Fig. \ref{fig:lrc_intro}(f)) with and without spacer (with a porosity of 83.8\% and spacers with 15\%). A zero porosity layer was added as a stiff substrate, such that when a vacuum is applied the structure would collapse in a bending motion. Similarly, a twisting actuator was printed with the spacers at a 45-degree angle to make a twisting motion the preferred deformation. Lastly, a screw pattern (see Fig. \ref{fig:lrc_intro}(f) with high porosity of 81\% and spacers of 15\%) had two zero-porosity sections perpendicular to each other. By spacing these two sections in succession, actuation would lead to a screw-like motion. The outer dimensions were set the same as the bending actuators. After printing, all printed actuators were put in a 0.4 mm thick SEBS (heat-sealed) sleeve and were actuated using a vacuum.

\begin{figure}[h]
 	\centering
 	\includegraphics[width=0.45\textwidth]{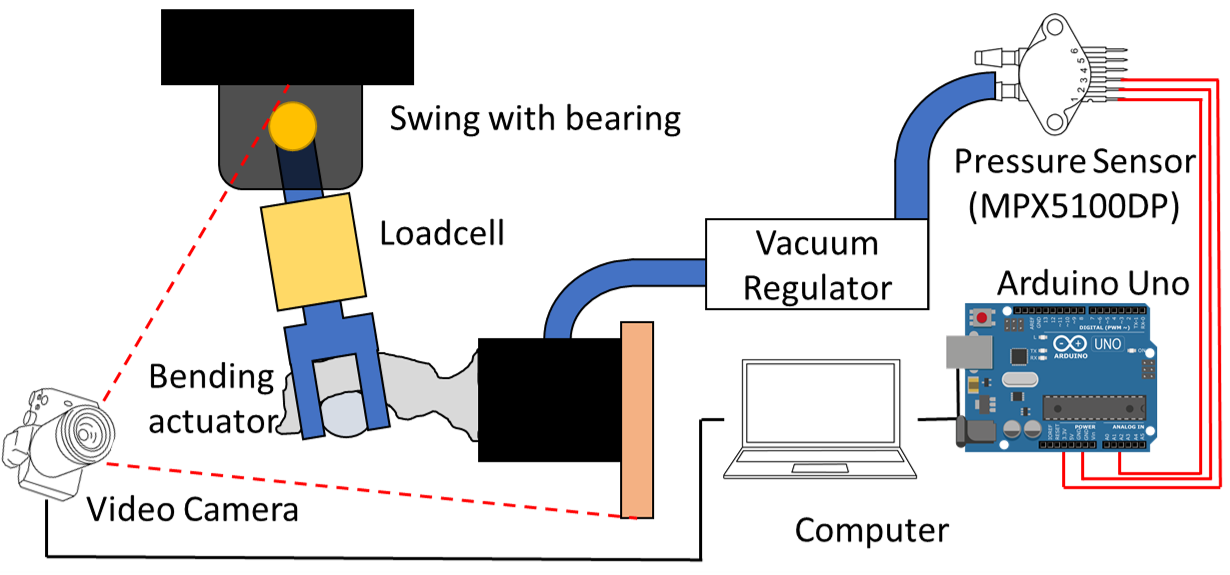}
 	\caption{Actuator characterization setup.}
 	\label{fig:ba1}
\end{figure}

The bending actuators with spacers were further investigated to characterize the effect of porosity on the output force and stroke. To this end, four soft bending actuators were printed with different levels of porosity ($\phi=[77.6,79.1,83.8,85.2]$). The nominal dimensions of these actuators were 15 mm wide, 75 mm long and 8 mm high. These actuators used a zero porosity strip as a (relatively) stiff substrate (1 mm thick). Subsequently, the actuators were put in a heat-sealed SEBS sleeve (0.4 mm thick). Both the sleeves and actuators were fabricated in triplicate.
The setup used for characterization of the output force and stroke evaluation is shown in Fig. \ref{fig:ba1}. The swing acts as a grasping object with a bearing to reduce friction. An LSB200 loadcell (Futek, USA) was integrated into the swing to measure the output force, which was connected to an Arduino Uno (Arduino AG, Italy) through an HX711 loadcell amplifier (Avia Semiconductor, China). The loadcell was tared right before starting a measurement to offset the mass of the actuator. For actuation, a vacuum source was operated using a manually triggered valve, and the vacuum pressure was set using a flow valve. The pressure was read by the Arduino Uno using an MXP5011DP pressure sensor (NXP Semiconductors, The Netherlands). The Arduino and a USB webcam (to record the actuator's deformation) were connected through USB to MATLAB. This experiment was performed in triplicate for each actuator for 30 seconds.

For analysis of the data, a Maxwell viscoelastic model\cite{Macosko} with one term was fitted to the force data, which is defined as:
\begin{equation}
	F(t) = (K_0 + K_1 \exp(-t/\tau_1))F_{max}
\end{equation}
Wherein $t$ is the time in seconds after the maximum force $F_{max}$ with $K_0$ and $K_1$ fitting parameters and $\tau_1$ the relaxation time. This equation, was used to predict the steady-state force ($F_{ss}=K_0F_{max}$) and settling time. The settling time was set to the time needed to be within 5\% of final value (i.e. $0.05K_0 = K_1\exp(-t_s/\tau_1)$).

In addition, the bending actuators were characterized in terms of stroke using the same setup. For these experiments (repeated thrice), the swing was removed and the actuator pressurized. Afterwards, the captured images were used to compute the stroke. Specifically, MATLAB was used to fit a circle along the deformed structure using three points to compute the maximum curvature. 

%%%%%%%%%%%%%%%%%%%%%%
\section{Results and Discussion}
 
\subsection{Characterization of Coiling Pattern}
\begin{figure*}[h]
 	\centering
 	\includegraphics[width=1.0\textwidth]{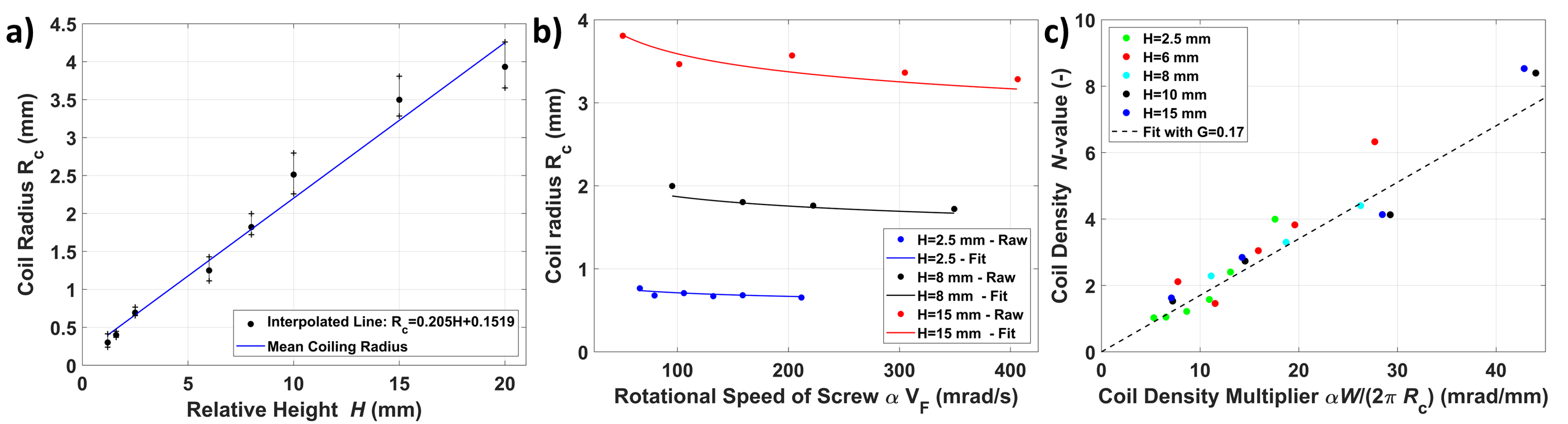}
 	\caption{(a,b) Coil radius versus the height $H$ (a) and (b) rotational speed of screw ($\alpha V_F$), and (c) The coil density ($N$-value) versus coil density multiplier $(\alpha W(2\pi R_c))$ used to estimate $G$ (see Equation \ref{eq:N}).}
 	\label{fig:RC}
\end{figure*}

The effect of the height ($H$) on the coil radius ($R_c$) is shown in Fig. \ref{fig:RC}(a). The plotted values of $R_c$ are plotted for multiple values of screw rotational speed ($\alpha V_f$ with $V_f$ the printhead speed) at the same height. It can be observed that $R_c$ is linearly related to the height ($H$) in the range of 2.5 mm to 15 mm, with almost one order of magnitude of increase. In contrast, the increase of $R_c$ seems to stall between 15 and 20 mm. This stalling implies that increasing $H$ will no longer increase $R_c$ (and could even decrease)\cite{ribe2012liquid}. This behavior makes further increasing the height not interesting for the InFoam method as it does not allow for larger coils. In addition, it further cools down the thermoplastic, which can lead to bonding issues making it even less appealing.

Besides $H$, the literature indicates that $R_c$ decreases when the viscosity decreases\cite{ribe2012liquid}. Due to the shear-thinning behavior of the thermoplastic, a decrease in viscosity (for constant temperature) should therefore coincide with an increased extrusion speed. The extrusion speed is equivalent (under the assumption of linearity) to an increased rotational speed of the screw $\alpha V_F$, which is plotted against $R_c$ in Fig. \ref{fig:RC}(b). It can be observed that there is a nonlinear and downwards trend with increasing extrusion speed. This behavior is in line with a power-law shear-thinning relationship. To compare the shear-thinning behavior, the viscosity of the SEBS material was first measured using a Rheocompass MC501 (Anton-Paar GmbH, Germany) with the data shown in Fig. S2 of the Supplementary Information. The data were then fitted to a power-law normalized by $R_c$ (using the exponent of the viscosity ($n-1=-0.09$), average coil radius $\hat{R}_c$ per $H$, and rotational speed of the screw $\alpha V_F$):
\begin{equation}
    R_c = a (\alpha V_F)^{n-1} \hat{R}_c
\end{equation}
Fitting the value of $a$ for all heights and averaging leads to the curve seen in Fig. \ref{fig:RC}(b). The fits had errors of less than 10\%. Thereby implying that there is a correlation between the change in coiling radius $R_c$ and the extrusion speed, which can explain the spread of coiling radii. This extrusion speed-induced change provides an additional opportunity to fine-tune $R_c$ at a fixed $H$. However, increasing the extrusion speed will also affect other properties such as the diameter (by increasing die swell).

The $N$-value was plotted against the coil density multiplier $\alpha W/(2 \pi R_c)$ (see Equation \ref{eq:N}) in Fig. \ref{fig:RC}(c) (using the mean value of $R_c$ per $H$). It can be observed that there is indeed a linear relation for the $N$-value. The linearity is also observed by fitting a line with $G=0.17$ mm/mrad (Equation \ref{eq:N}). The accuracy thereof implies that the proposed relationship is valid. In addition, when extruding in the air the value of $G$ was found to be in the range of 0.154-0.165 mm/mrad, which is in the same range as the fitted value (see Supplementary Information). Lastly, it can be observed that lower $H$ do not achieve high $N$-values. This discrepancy is due to the coils stacking up too much. When this occurs the coiling pattern is replaced with an accumulation of material\cite{yuk2018new}. This accumulation phase is more chaotic and not used by the InFoam method.  

Lastly, the coil height ($h_c$) estimation (Equation \ref{eq:hc}) was validated using the coiling patterns to be on average within 4\% (see Supplementary Information).

\subsection{Mechanical Programming of Soft Cubes}
\begin{figure*}[h]
 	\centering
 	\includegraphics[width=1.0\textwidth]{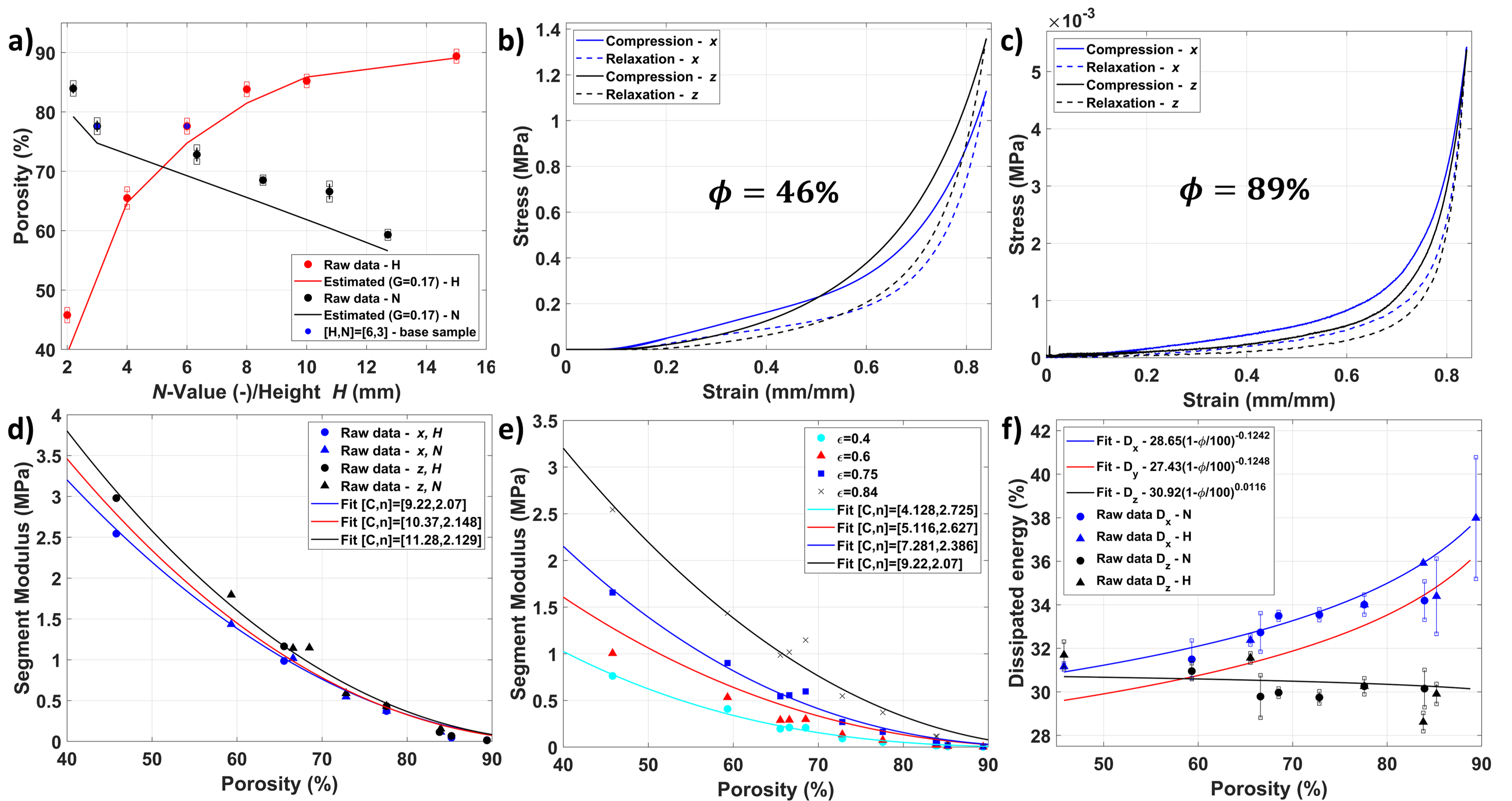}
 	\caption{(a) Porosity versus height ($H$) and coil density ($N$-value), (b,c) stress-strain curve at 46\% and 89\% porosity, (d,e) segment modulus versus porosity for different axes at maximum strain (d) and different levels of strain (e), and (f) the energy dissipation versus porosity.}
 	\label{fig:poroest2602}
 \end{figure*}

The measured and computed porosity of the soft cubes versus the height ($H$) and the coil density ($N$-value) are shown in Fig. \ref{fig:poroest2602}(a). It can be observed that the porosity ranged from 45 to 89\% with similar levels of porosities were attained by changing $H$ and the $N$-value. Increases in the $N$-value/$H$ lead to a linear decreasing and nonlinear increase of the porosity, respectively. The predicted porosity of Equation \ref{eq:Gman} (with the estimate of $G$ from the previous section) correlated quite well. Even though the accuracy for the $N$-value is lower than the coil radius ($R_c$) estimate, the mean absolute error was less than 4\% for both ($H$ (2.6\%) and the $N$-value 3.9\%). This graph demonstrates that the $N$-value can be used to tune the level of porosity independent of $H$. This change in porosity is advantageous as it means the InFoam method can adjust the $N$-value by setting the extrusion multiplier ($\alpha$) accordingly (see Equation \ref{eq:N}). By doing so the InFoam method can use both large and small coiling radii to increase accuracy (smaller) or printing speed (larger), while still achieving the desired porosity. 

The results of the compression tests on the cubes were analyzed with the axes defined as: in line with the coils $x$, perpendicular to the coils but in-plane $y$, and layer-wise $z$. The stress-strain behavior of the soft cubes are shown in Fig. \ref{fig:poroest2602}(c) and (d) for the $x$- and $z$-axis ($y$-axis is similar but not shown for clarity) for both low (45.8\%,(c)) and high (89\%,(d)) porosity, respectively. Their general shape is similar with a clear hysteretic behavior but with much lower magnitudes for the higher porosity cube. It can also be noted that both have a linear region and a nonlinear region. The nonlinearity occurs due to the densification of the porous structure (i.e. when all air has been pushed out). This densification limits the effective deformation one can practically achieve\cite{mac2018compliant}, which is reached at an earlier stage in the higher porosity than the lower porosity. The stress in $x,z$ seems linear up to 0.2 and 0.5 strain, for the high and low porosity cube, respectively. The compression of three foam cubes with different porosity is shown in Supplementary Movie 2 in the Supplementary Information. 

The segment modulus at maximum strain (0.84) is plotted in Fig. \ref{fig:poroest2602}(d) for all three axes. The raw data is shown for the $x$ and $z$ axis (see Fig. S3 of the Supplementary Information for the $y$). Error bars were not added for clarity but the standard errors were on average within 13\%.  It can be observed that the trend in both $N$-value and $H$ are similar. This observation implies that the magnitude of the porosity is important and not the geometry of the underlying coiling pattern (i.e. the specific $R_c$ and $N$-value combination). In addition, it can be observed that all three axes have a power-law relation with the porosity, as expected.  The fitted line shows that a (nearly) quadratic function describes all three directions reasonably well, which is in line with the decay of the elastic modulus seen in normal foams\cite{gibson1982mechanics}. In addition, different moduli can be observed for the three axes, which indicates that the layout of the coiling pattern will induce some anisotropy. A possible method to reduce the anisotropy between $x$ and $y$ is to change the coil trajectory by 90 degrees between layers. This approach leads to similar stress in the $x$ and $y$-direction (see Fig. S4 in the Supplementary Information).

The segment modulus was also evaluated at multiple levels of strain for the $x$-axis (see Fig. \ref{fig:poroest2602}(e)). It can be observed that before densification (low strain) the moduli are more similar. After densification the slope at which moduli changes increases significantly (from 11.25 to 19.09). Interestingly, the modulus decreases faster but still in a very similar magnitude to normal foams (i.e. around quadratically)\cite{gibson1982mechanics}. This change allows for mechanical programming to have parts that behave very differently under the same load. To compare the magnitudes to the bulk material, the modulus was approximated using the Shore hardness (47A) and the equation\cite{Dow} $E=0.486\exp{(0.0345\cdot 47)}=2.46$ MPa. It can be observed that even at a strain of 0.4 that the 45.8\% and 89\% cubes were 3.26 and 246 times as small. Thereby indicating that the InFoam method can reduce the modulus by over two orders of magnitude. This value exceeds the 66 times already demonstrated in literature when printing with LRC\cite{lipton20163d}.

Lastly, the dissipated energy (in percentage) versus porosity is plotted in Fig. \ref{fig:poroest2602}(f) with the standard deviation. It can be observed that the $x$-direction (with 31-40\%) dissipates the most energy. A power-law relationship with the porosity can be observed for the $x$ and $y$ directions but is nearly constant in $z$. We expect that is due to the buckling of coils in both $x$ and $y$, whereas the coils are compressed/bent in the $z$-direction. In addition, the effect of $H$ and the $N$-value are similar (just as for the stiffness) providing more evidence that not the coil geometry but porosity is the important value.

\subsection{Graded Porosity for Soft Fluidic Actuators}
The printed structures with porosity gradients are shown in Fig. \ref{fig:porofluid2} and Supplementary Movie 3 (see Supplementary Information). It can be seen that the porosity gradient can deform to the screw motion, which is shown without a sleeve (Fig. \ref{fig:porofluid2}(a)) and actuated (\ref{fig:porofluid2}(b)). In addition, the twisting (\ref{fig:porofluid2}(c)) and contraction (\ref{fig:porofluid2}(d)) are shown in both actuated and unactuated form. Lastly, Fig. \ref{fig:porofluid2}(e) and (f) show that the absence (\ref{fig:porofluid2}(e)) and presence of a porosity gradient with spacers (\ref{fig:porofluid2}(f)) can significantly increase the bending. In general, screw, twisting, and two bending actuators all have similar external geometry but using different porosity gradients allows for a wide variety of deformation patterns. These examples show that the InFoam method can change the deformation behavior significantly. This grading capability during AM enables designers to design the deformation behavior purely by the porosity gradient.

\begin{figure}[h]
 	\centering
 	\includegraphics[width=0.5\textwidth]{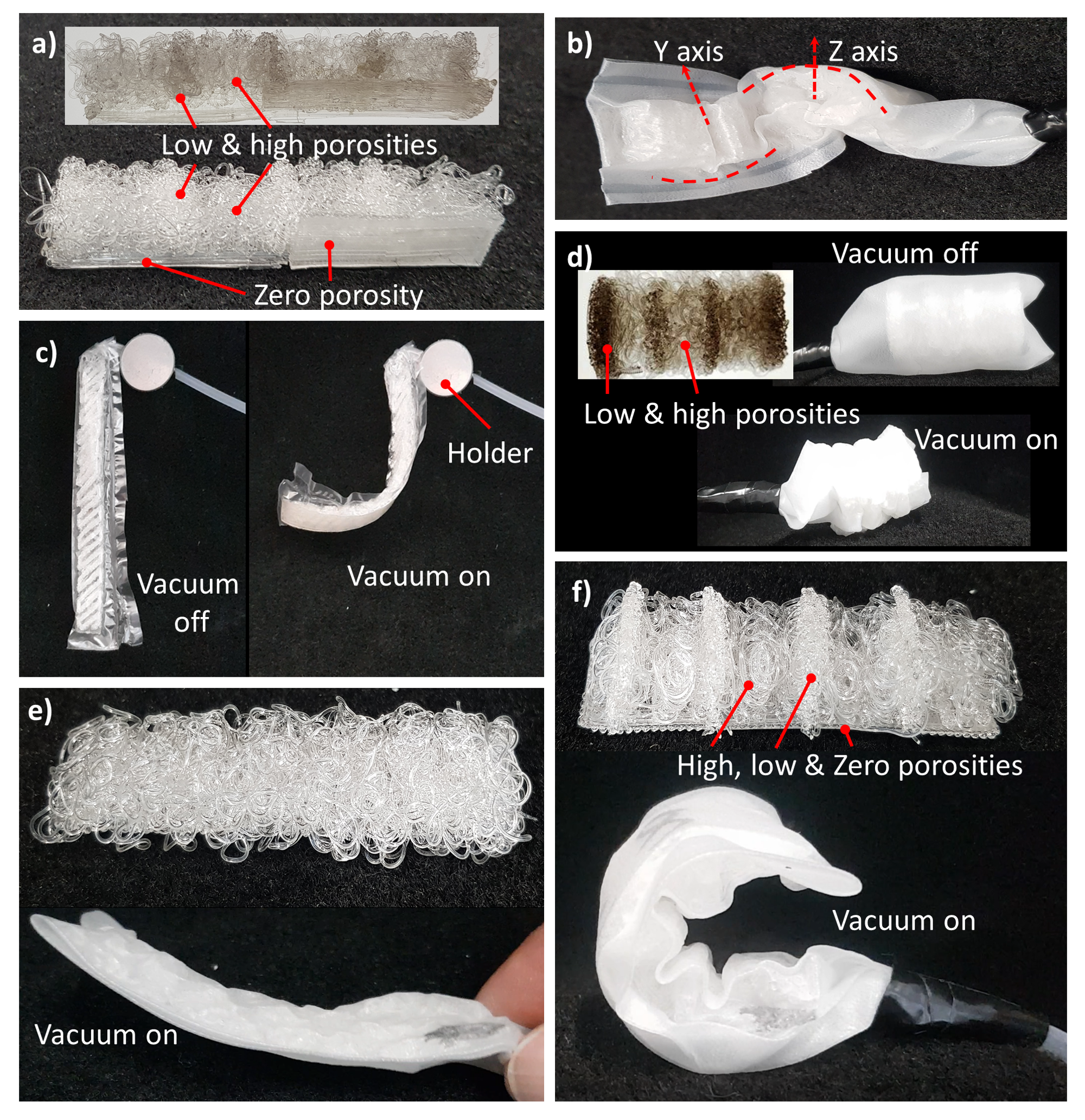}
 	\caption{Deformation pattern for different porosity patterns.}
 	\label{fig:porofluid2}
 \end{figure}
 
 \begin{figure*}[!b]
 	\centering
 	\includegraphics[width=1.0\textwidth]{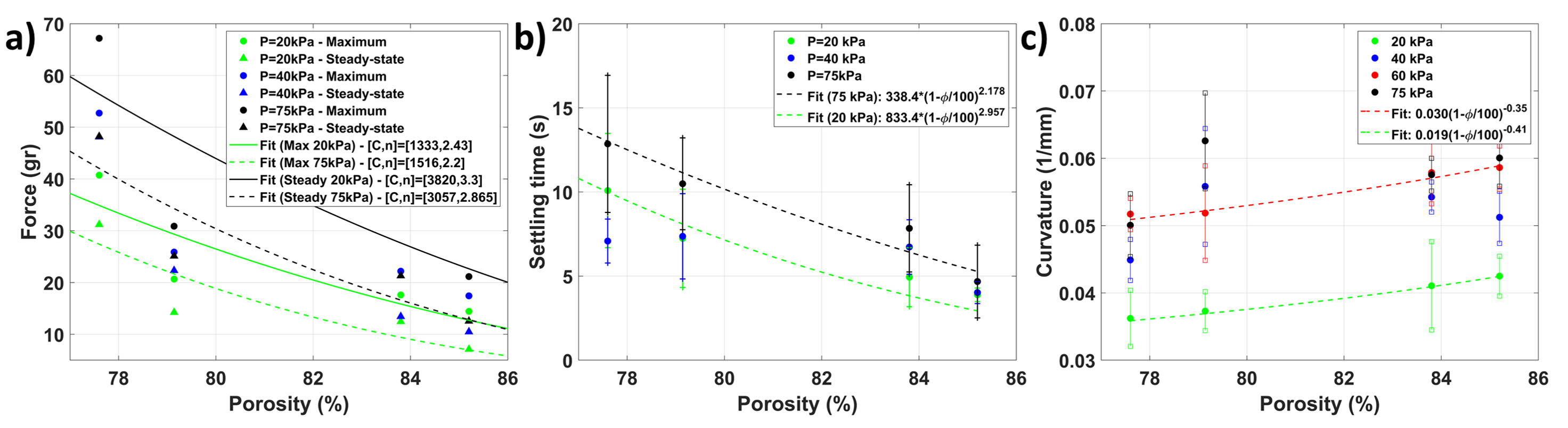}
 	\caption{(a) Maximum and steady state force, (b) settling time, and (c) curvature.}
 	\label{fig:ba2}
 \end{figure*}
 
The maximum and steady-state force of the bending actuators (of Figure \ref{fig:porofluid2}(f)) are plotted in Fig. \ref{fig:ba2}(a) (and  in Supplementary Movie 4 of the Supplementary Information). For clarity of the image, the error bars are omitted but these were lower than 10\% for most (the data is provided with error bars in Fig. S5 in the Supplementary Information). It can be observed that an increase in pressure leads to an increase in both steady-state and maximum force. The maximum forces were in the range of 67.2 to 21.1 grams for actuators, which is 10.2 to 3.8 times their weight (6.6 and 5.5 grams, respectively). In addition, a decrease in force for higher porosity actuators is observed. This behavior is expected to be due to the lower stiffness. The lower stiffness would allow the actuator to deform such that the output force is reduced. The steady-state value of the force increases less with increasing pressure for lower porosity samples. This trend is expected to be linked to densification, which decreases the achievable deformation of lower porosity samples. Fitting a curve for the maximum and steady-state force curve gives exponents that are similar to each other. However, there is a discrepancy between the maximum and steady-state values in terms of their exponent. This discrepancy is expected to be due to the viscous component of the thermoplastic, which impacts the maximum force only. Interestingly, the power-law fits for both maximum and steady-state force correlate well with data, which indicates that porosity could be an interesting metric to predict the mechanical performance of an SFA. 

The settling time, i.e. when the force is within 5\% of the steady-state value, is plotted in Fig. \ref{fig:ba2}(b). Although there is a significant spread, the trend does indicate that the settling time decreases with increasing porosity. This decrease is expected to be due to the combination of decreased density and increased energy dissipation associated with higher porosity, as seen in the soft cubes. Fits based on the average settling time for 20 and 75 kPa, follow a power-law behavior with the porosity. Thereby supporting that the porosity could be an interesting metric to predict the mechanical performance. 

Lastly, the maximum curvature versus porosity for different pressures is shown in Fig. \ref{fig:ba2}(c). A nonlinear relation can be observed between the curvature and porosity. This behavior is expected as the higher porosity actuators will deform more under the same load. Similar to the force metrics discussed previously, these curves can be described with a power-law relation. Fitting the data shows that the curvature increases with the porosity with an exponent around the cubic root of the porosity. Thereby indicating again that porosity could be a useful tool for programming deformation behavior. In addition, a nonlinear relation between the curvature and pressure can be observed. Independent of the porosity the actuators seem to plateau at 60 and 75 kPa. This plateau is expected to be due to the densification of the porous structure. The densification means that exponentially more pressure would be required to increase curvature. The higher densification strain of the higher porosity structures allows them to deform more before reaching this point. 

%%%%%%%%%%%%%%%%%%%%%%%%%%%%%%%%%%%%%%%%%%%%%%%%%%%%%%%%%%%%%%%%%%%%%%%%%%%%%%%%%%%%%%%%%%%%%%%%%%%%%%%%%%%%%%%%%%%%%%%%%%%%%%%%%%%%%%%%%%%%%%%%%%%%%%%%%%%%%%%%%%%%%%%%%%%%%%%%%%%%%%%%%%%%%%%%%%%%%%%%%%%%%%%%%%%%%%%%%%%%%%%%%%%%%%%%%%%%%%%%%%%%%%%%%%%%%%%%%%%%%%%%%%%%%%%%%%%%%%%%%%%%%%%%%%%%%%%%%%%%%%%%%%%%%%%%%%%%%%%%%%%%%%%%%%%%%%%%%%%%
\section{Conclusion}
Soft fluidic actuators are a popular choice in soft robotics due to their versatility, accessibility, and (relative) simplicity. To further expand their capabilities, fabrication methods for structures that simultaneously allow for the transfer of fluidic power and have a stiffness gradient are required. Our InFoam method, as proposed in this work, can provide this capability by allowing for the direct printing of porous structures. The printed porous structures allow for fluid transport and mechanical flexibility. In addition, the structures used a hyperelastic thermoplastic to further reduce the stiffness due to the usage of a hyperelastic thermoplastic.

The InFoam method is capable of mechanical programming by a single level or gradient of porosity (by combining it with normal plotting). To achieve this capability it only requires the characterization of the coil radius (with respect to the height) and the length of extruded material per unit of movement ($G$) at the printing temperature. Subsequently, the InFoam method can use the developed equations to adjust the porosity of a structure by setting the height and/or extrusion speed while taking the coiling pattern's width and height into account.

With a single level of porosity, the InFoam method can program the density, stiffness, and energy dissipation. Large changes could be realized such as an 89\% lower density and scaling the modulus by a factor of 246. In addition, the energy dissipation in $x$ and $y$ could be programmed without affecting the $z$-direction. All of these properties were shown to correlate well through a power-law relationship with the porosity. 

By printing a porosity gradient, the InFoam method allowed for a wide range of motion (including twisting, contraction, and bending). In addition, the grading was shown to be capable of programming the behavior of soft bending actuators. Specifically, it was shown that the porosity could be used to program both the output force (magnitude and settling time) and stroke.

Thus, the InFoam method can print graded porous structures that are mechanically programmed by porosity. The InFoam method allows this capability by solely adjusting the height of the nozzle and the ratio of extrusion and movement speed. An interesting observation of the printed porous structures is that the mechanical behavior of both the cubes and bending actuators could be correlated through a power-law relationship with the porosity. In addition, they showed similar trends in their changes in behavior. Further investigation of this correlation could enable tuning of the mechanical performance of soft (porous) actuators using a simple compression experiment.
In addition, the InFoam method's stiffness grading can be further investigated. A structure with high porosity has low stiffness and allows fluid transport, which can be interesting for applications such as scaffolds for more complex SFAs and support. Whereas graded porosity can be a tool for soft fluidic actuators/sensors with complex deformation patterns.

\bibliographystyle{IEEEtran}
\bibliography{template}  

\end{document}